\documentclass{article}

\usepackage[utf8]{inputenc} 
\usepackage[T1]{fontenc}    
\usepackage{url}            
\usepackage{booktabs}       
\usepackage{amsfonts}       
\usepackage{nicefrac}       
\usepackage{dsfont}
\usepackage{microtype}      
\usepackage{lipsum}
\usepackage{fancyhdr}       
\usepackage{graphicx}       
\graphicspath{{media/}}     
\usepackage{amsmath}
\usepackage{amssymb}
\usepackage{bm}
\usepackage{algorithm, algorithmic}
\usepackage{cite}
\usepackage{amsthm}

\usepackage{hyperref}       
\newtheorem{theorem}{Theorem}[section]

\title{Reinforcement Learning with Prior Policy Guidance for Motion Planning of Dual-Arm Free-Floating Space Robot}

\author{
    Yuxue Cao$^1$, Shengjie Wang$^{2*}$, Xiang Zheng$^3$, Wenke Ma$^4$,\\ Xinru Xie$^1$, Lei Liu$^{1*}$\\
    \small $^1$Beijing Institute of Control Engineering \\ 
    \small $^2$ Department of Automation, Tsinghua University \\
    \small $^3$ Department of Computer Science, City University of Hong Kong \\
    \small $^4$ Qian Xuesen Laboratory of Space Technology \\
}
\date{}

\begin{document}
\maketitle

\begin{abstract}
\noindent Reinforcement learning methods as a promising technique have achieved superior results in the motion planning of free-floating space robots. However, due to the increase in planning dimension and the intensification of system dynamics coupling, the motion planning of dual-arm free-floating space robots remains an open challenge. In particular, the current study cannot handle the task of capturing a non-cooperative object due to the lack of the pose constraint of the end-effectors. To address the problem, we propose a novel algorithm, EfficientLPT, to facilitate RL-based methods to improve planning accuracy efficiently. Our core contributions are constructing a mixed policy with prior knowledge guidance and introducing $\| \cdot \|_{\infty}$ to build a more reasonable reward function. Furthermore, our method successfully captures a rotating object with different spinning speeds. \footnote{\textit{\underline{Corresponding author}}:\textbf{Lei Liu}(liulei@spacechina.com), \textbf{Shengjie Wang}\\(wangsj19@mails.tsinghua.edu.cn)}

\end{abstract}

\section{Introduction}
In the long term, there exists the failure of satellites in the complex and variable space environment due to aging, debris impact, or fuel depletion. Apparently, it can significantly reduce the risk and cost by replacing astronauts with space robots to carry out on-orbit services to save the satellites' life \cite{li2019orbit, sellmaier2010orbit, flores2014review}. For the on-orbit service, an essential task is to plan and control the manipulator's motion so that the end-effector can reach the target along a collision-free and non-singular path. Compared with the single-arm space robot, the dual-arm system has expanded workspace, higher flexibility, and better load-carrying capacity. However, the dual-arm cooperative planning and control problem of free-floating space robots remains a critical challenge.

\noindent For the planning method in Cartesian space and then obtaining the corresponding joint trajectory by the generalized Jacobian matrix (GJM) \cite{papadopoulos1993dynamic, umetani1989resolved, yoshida1991dual }, it is necessary to balance the contradiction between avoiding dynamic singularity and improving the tracking accuracy. To solve the dynamic coupling problem, researchers used the reactive zero-space algorithm \cite{zhou2017singularity, lu2020multi} to reduce the interference of the base, thus improving tracking accuracy. However, it is impossible to avoid the complex modeling and calculation process. In transforming the planning problem into the optimization problem \cite{wang2018optimal,zhang2018effective,zhao2019multitask,wang2018coordinated,yan2020multi}, the traditional optimization methods only retain the optimal samples, thus having low sampling efficiency and slow iterative convergence. Heuristic algorithms such as RRT \cite{xie2019obstacle, zhang2020sampling} are suitable for planning with a determined start and target configuration. When the manipulator's initial configuration or the target object changes, the trajectory must be recalculated. Furthermore, after obtaining the desired trajectory in Cartesian or joint space, it is often necessary to adjust the controller manually to ensure the tracking accuracy \cite{liu2021trajectory}.

\noindent In recent years, model-free reinforcement learning has rapidly developed and been applied to robot planning and control \cite{yamada2020motion, xia2021relmogen, elhaki2021novel, he2021explainable}. It relies on the interaction between the agent and the environment to acquire the corresponding ability, thus avoiding complex modeling and manually adjusting controller parameters. Classical deep reinforcement learning algorithms such as PPO \cite{wang2021multi}, DDPG \cite{du2019learning,hu2018mrddpg} and SAC \cite{wang2021end} have been proved to be adequate to solve the planning problem of the free-floating space robots. However, the existing research mainly realized the position and orientation planning for the single-arm free-floating space robots. For the collaborative planning of dual-arm space robots, simultaneously considering the position and orientation of both arms will significantly exacerbate the optimization difficulty. To the best of our knowledge, the previous studies only realized the motion planning of the end-effectors' positions. Due to the lack of the constraint of the end-effectors' poses, such methods can not solve the task of capturing a spinning object (non-cooperative object).

\noindent To address the issue of low efficiency, we propose the \textbf{EfficientLPT} (Efficient Learning-based Path Tracking) algorithm for simultaneous motion planning of the position and orientation of the dual-arm free-floating space robot. More importantly, our method performs the task successfully, tracking multiple targets' poses and capturing a spinning target by a dual-arm free-floating space robot. 
We further made the following contributions:
\begin{itemize}
\item We introduce the inverse kinematics of the fixed base manipulator as a prior policy to guide the agent's explorable direction close to the optimal policy. We illustrate that the mixed policy with prior knowledge guidance solves the cold-start problem and greatly improves the sample efficiency in theoretical and experimental views.

\item We build a novel reward function related to reducing the end-effector's orientation error by only considering the infinite norm of the axis angle error, thus expediting the algorithm's convergence.

\item Thanks to the efficiency and robustness of \textbf{EfficientLPT}, we realize the motion planning and tracking of the target rotating at different speeds without further training.

\end{itemize}

\section{Related Work}
Many researchers used model-free methods in motion planning to avoid dynamic singularities of the free-floating space robots. Yan \cite{yan2020multi} and Wang \cite{wang2018coordinated} respectively used the fifth-order polynomial and Bézier curve to parameterize the joint trajectories, then used the PSO algorithm and the differential evolution algorithm to find the optimal trajectory under specific objectives and constraints. Zhao \cite{zhao2019multitask} employed piecewise continuous-sine functions to depict the joint trajectories and improved the genetic algorithm to optimize the parameters. When the planning dimension increases, the iterative convergence efficiency of these traditional optimization methods is low. Furthermore, when the trajectory starting or ending point changes, it has to find the optimal path again. Thus these algorithms have poor generalization.

\noindent In comparison, reinforcement learning has higher optimization efficiency and generalization, so it has become popular in manipulator planning. Zhong \cite{zhong2022collision} and Xie \cite{xie2019deep} verified the effectiveness of DDPG in manipulator planning and accelerated the algorithm convergence by introducing the manipulator inverse kinematics and a properly designed reward function, respectively. Reinforcement learning has also been applied to multi-arm cooperative planning. Prianto \cite{prianto2020path} performed dual-arm motion planning in the joint space based on the SAC algorithm. Ha \cite{ha2020learning} used the multi-agent RL algorithm to train a decentralized policy to control each manipulator to reach the target pose.

\noindent However, due to the dynamic coupling between the base satellite and the manipulator, the results obtained by previous work can not apply to some space operations directly. In the current literature, Du \cite{du2019learning} and Hu \cite{hu2018mrddpg} applied DDPG to the single-arm free-floating space robot planning problem. Wu \cite{wu2020reinforcement} and Li \cite{li2021constrained} further used DDPG for the collaborative planning of the dual-arm space robot. The former realized the tracking of targets moving at different speeds, while the latter considered tracking the target rotating around a single axis at a fixed speed. Nevertheless, the above researches only consider position planning, i.e., reducing the position error between the end-effector and the target. Meanwhile, to realize the capture tasks, it is also necessary to conduct orientation planning. Li \cite{li2022constrained} combined the artificial potential field algorithm (APF) with the reward function to solve the problem of falling into the minimum value in the APF. Wang \cite{wang2022collision} decoupled the manipulator planning into two sub-goals of position and orientation and realized the simultaneous planning of the position and orientation trajectory through the alternate training of two agents. But all these methods are only applied to the single-arm space robot. In this work, we simultaneously consider the position and orientation planning of the dual-arm free-floating space robot to track the target rotating at different speeds within a specific range, which is a further study on existing research.

\section{The Kinematic Model of Dual-Arm Free-Floating Space Robot}

\begin{figure}[H]
  \centering
  \includegraphics[width=\hsize]{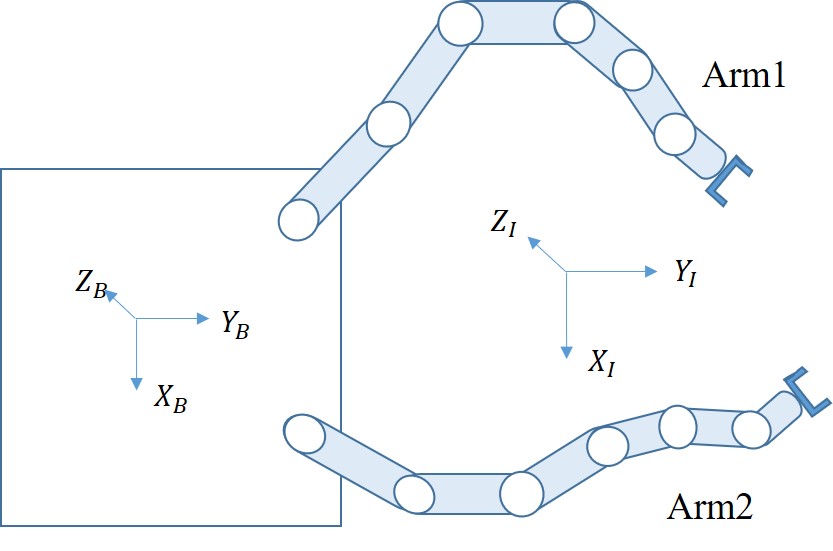}
  \caption{Model of the dual-arm free-floating space robot.}
  \label{Kmodel}
\end{figure}

The dual-arm free-floating space robot comprises a base satellite and two n-Dof manipulators. We assume that the base satellite, the manipulator links, and the joints connecting the components are all rigid bodies. In this paper, the base satellite is equipped with two 6-Dof UR5 manipulators whose kinematics and dynamics parameters are the official default values, as shown in Fig. \ref{Kmodel}.

\noindent The velocity of each manipulator's end-effector is
\begin{equation}
\label{eq1}
\dot x^i_{e}=J_{b} \dot {x_{b}}+J^i_{r} \dot{\Theta}_{i}
\end{equation}
where $J_b$ is the Jacobian matrix to the base satellite, $\dot x_{b}$ is the velocity of the base satellite, $J^i_r$  is the Jacobian matrix of Arm-$i$, and $\dot{\Theta}_{i}$ is the joint angular velocities of Arm-$i$.

\noindent Assuming the initial linear and angular momentum of the system is zero, the momentum conservation of the space robot can be formulated as
\begin{equation}
  \label{eq2}
H_{b}\dot {r_{b}}+H^1_{r} \dot{\Theta}_{1}+ H^2_{r} \dot{\Theta}_{2}=0
\end{equation}
where $H_b$, $H^1_r$, and $H^2_r$ are the coupling inertia matrices of the base, Arm-1, and Arm-2, respectively. 

\noindent According to Eq. \ref{eq2}, the base satellite velocity can be deduced as
\begin{equation}
\label{eq3}
\dot x_{b} = J_{a}\dot{\Theta} = 
\begin{bmatrix} 
{-H^{-1}_{b}H^1_r} & {-H^{-1}_{b}H^2_r} 
\end{bmatrix}
\begin{bmatrix} {\ddot{\Theta}_{1}} \\{\ddot{\Theta}_{2}} \end{bmatrix}
\end{equation}

\noindent Substituting Eq. \ref{eq3} into Eq. \ref{eq1}, the kinematic model of the dual-arm free-floating space robot is obtained as
\begin{equation}
\label{eq4}
\dot x^i_{e} = J_{g}\dot{\Theta} = 
\begin{bmatrix} 
{J^1_{r} - J_{b} H^{-1}_{b} H^1_{r}} & {-J_{b} H^{-1}_{b} H^2_r} \\
{-J_{b} H^{-1}_{b} H^1_r} & {J^2_{r} - J_{b} H^{-1}_{b} H^2_r}
\end{bmatrix}
\begin{bmatrix} {\dot{\Theta}_{1}} \\{\dot{\Theta}_{2}} \end{bmatrix}
\end{equation}

\noindent We can see that the GJM(Generalized Jacobian Matrix) $J_{g}$ is related to the system's inherent parameters and motion quantities. The trajectory of the end-effectors in Cartesian space is related to the coupling between the manipulators and the floating base when reaching and tracking the targets. Therefore, achieving accurate planning is a complex problem.

\section{Method}
\subsection{Planning Problems with Reinforcement Learning}
In capturing the target with two cooperative manipulators, the primary goal is to enable the end-effectors to accurately reach the target position and orientation. Secondly, the joint angles and joint angular velocities shall not exceed limitations during the movement. Moreover, since collision can easily lead to instability of the free-floating system, planning a collision-free path is also an essential requirement. Comprehensively considering the above requirements, the theoretical constraints can be described as:
\begin{equation}
\begin{cases}
    \Vert e_{p}(t_{f}) \Vert \leq U_{e_{p}} \\
    \Vert e_{o}(t_{f}) \Vert \leq U_{e_{o}} \\
    \Vert \theta_{i} \Vert \leq U_{\theta_{i}}, i = 1,...,6 \\    \Vert \dot{\theta_{i}} \Vert \leq U_{\dot{\theta_{i}}}, i = 1,...,6 \\
    S(p_{b}, q_b, \Theta) \cap O = \emptyset
\end{cases}    
\end{equation}
Each $U_{*}$ represents the upper limit of the corresponding parameter, including the position error $e_{p}$ and orientation error $e_{o}$ between the end-effectors and the target, the joint angles $\theta_{i}$, and the joint angular velocities $\dot{\theta_{i}}$. $S(p_{b}, q_b, \Theta)$ represents the whole space robot system, and O is the set of all obstacles in the workspace. The intersection of the two spaces is constrained to be $\emptyset$ means that the planning path is collision-free.

\noindent Usually, the degree of freedom of reaching constraints is used to quantify the difficulty of the planning task. The dual-arm collaborative planning problem discussed in this paper is a 12-dimensional complex problem. The path planning algorithm based on reinforcement learning is proposed. 

\begin{figure}[H]
  \centering
  \includegraphics[width=\hsize]{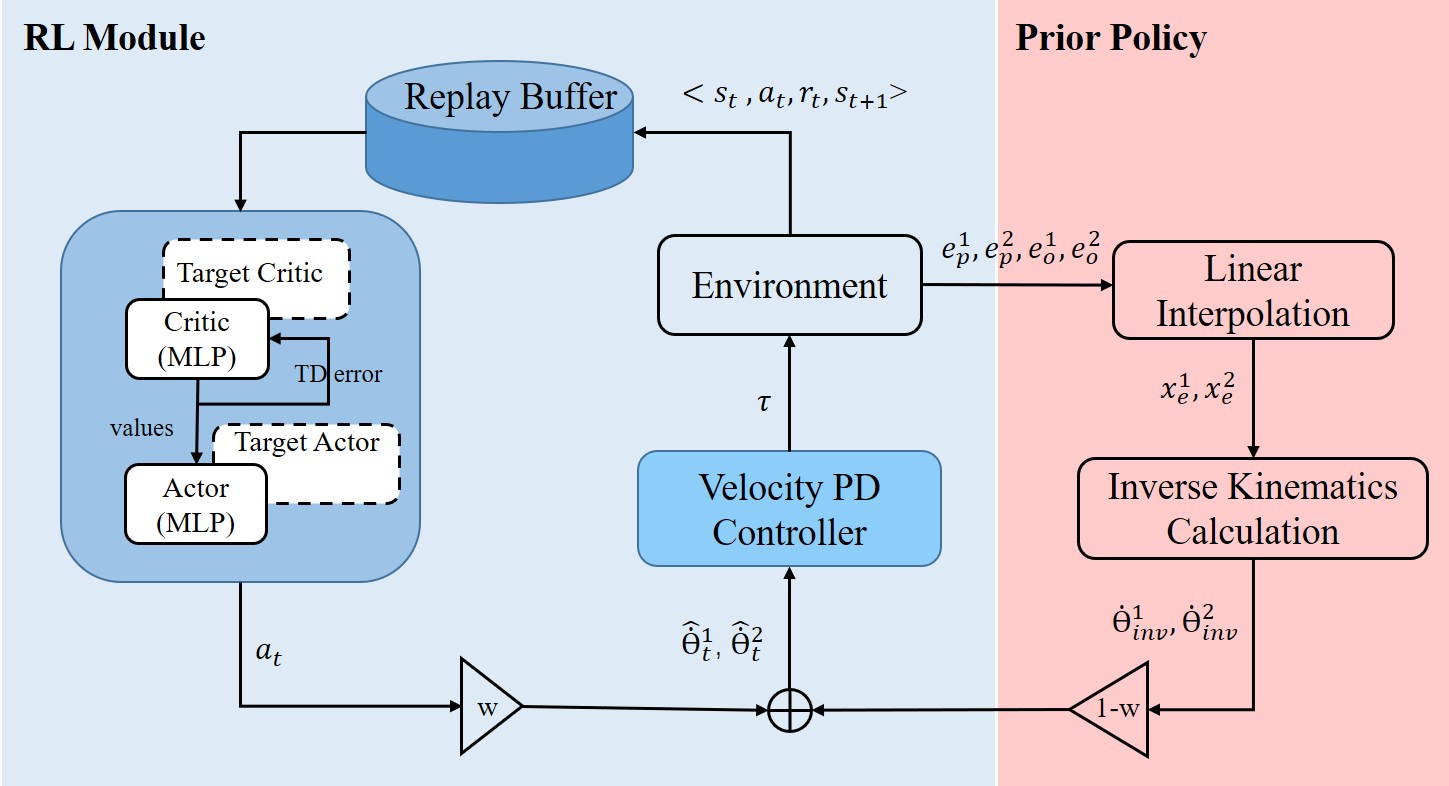}
  \caption{This is the framework of our algorithm. \textbf{RL Module} represents the SAC learning framework expounded in Section 4.2. \textbf{Prior Policy} is the prior knowledge guidance we introduced, which will be described in detail in Section 4.3.}
  \label{Algframework}
\end{figure}

\noindent For the task capturing a target with motion uncertainty, the agent obtains the states including the joint angles $\theta_{1,2} \in \mathbb{R}^6$, the joint angular velocities $\dot{\theta}_{1,2} \in \mathbb{R}^6$, the pose of end-effectors ${P_{{e}_{1,2}}, \Phi_{{e}_{1,2}}} \in \mathbb{R}^6$ of two manipulators and the pose of target points ${P_{{t}_{1,2}}, \Phi_{{t}_{1,2}}} \in \mathbb{R}^6$. In order to accelerate the algorithm's convergence, the position error vector $e_{{p}_{1,2}} \in \mathbb{R}^3$ and the three-axis angle error vector $e_{{o}_{1,2}} \in \mathbb{R}^3$ between of the end-effectors and the target are also taken into the state space. As for action space, the agent outputs the desired joint angular velocities to obtain smoother manipulator movement. In low-level planning, a PD controller converts the desired velocities into the torques, then controls the manipulators. Fig. \ref{Algframework} shows the framework of the planning algorithm proposed in this paper. The details will be elaborated on in the rest of this section.

\subsection{SAC Algorithm}
Unlike other reinforcement learning algorithms that only maximize the agent's reward, SAC \cite{haarnoja2018soft} utilizes entropy regularization to improve the policy exploration by maximizing its entropy, which is more effective for the multi-goal optimization problem. Therefore, the optimal policy of SAC maximizes both the reward obtained in each episode and the policy entropy, as shown in Eq. \ref{eq6}.

\begin{equation}
\label{eq6}
{\pi ^*} = \arg {\max _\pi }{{\rm{E}}_{({s_t},{a_t}) \sim {\rho _\pi }}}[\sum\nolimits_t {R({s_t},{a_t}) + \alpha H(\pi ( \cdot \left| {{s_t}} \right.))]} 
\end{equation}

\noindent SAC is a deep reinforcement learning algorithm based on the Actor-Critic framework. The Actor is the policy function to generate actions that the agent interacts with the environment. At the same time, the Critic is the value function responsible for evaluating the actions that Actor outputs to guide the policy update. Referring to the Bellman Equation, when maximizing the reward values, the policy entropy can be regarded as a part of it. Therefore, the updating of the value function of the Critic based on the time-difference (TD) method can be described as Eq. \ref{eq7}: 

\begin{equation}
\begin{split}
\begin{aligned}
\label{eq7}
{J_Q}(\theta) = {{\mathbb{E}}_{({s_t},{a_t},{s_{t + 1}}) \sim D,{a_{t + 1}} \sim {\pi _\phi }}}[\frac{1}{2}({Q_\theta }({s_t},{a_t}) - (r({s_t},{a_t}) + \\\gamma ({Q_{\bar \theta }}({s_{t + 1}},{a_{t + 1}}) - \alpha \log ({\pi _\phi }({a_{t + 1}}\left| {{s_{t + 1}}} \right.))))^2]
\end{aligned}
\end{split}
\end{equation}

\noindent Like DDPG, SAC contains a target soft Q-network $Q_{\bar \theta }$ whose parameters are obtained by Q-network exponentially moving average. In order to solve the Q-value overestimation in the update of the value function network, the algorithm introduces the double Q-learning \cite{van2016deep}, i.e., using two independent Q-learning networks $Q_{\theta_{1}}$ and $Q_{\theta_{2}}$. The update of the target Q-network is formulated as Eq. \ref{eq8}. The Q-value overestimation is alleviated by taking the minimum value of the two Q-learning networks. The policy network is updated to minimize the policy KL-divergence, as formulated in Eq. \ref{eq9}.
    
\begin{equation}
\label{eq8}
Q_{\bar \theta }(s_{t + 1},a_{t + 1}) = r_{t} + \mathop{\min }  \limits_{i = 1,2} Q_{\theta_{i}} (s_{t + 1},\mu_{\theta}(s_{t + 1}))
\end{equation}

\begin{equation}
\begin{split}
\begin{aligned}
\label{eq9}
{J_\pi }(\phi ) = {\mathbb{E}_{{s_t} \sim {\mathcal{D}}, {a_t} \sim \pi_{\phi}}}[\log {\pi _\phi }({a_t} \left| {{s_t}} \right.) - \frac{1}{\alpha}{Q_\theta }({s_t},{a_t })]
\end{aligned}
\end{split}
\end{equation}

\noindent As an off-policy reinforcement learning algorithm, SAC collects transition tuples $(s_{t}, a_{t}, s_{t+1}, R(s_{t}, a_{t}))$ during each episode of agent interaction with the environment, and stores them in the replay buffer. These samples are used to update the Actor and Critic networks during training.

\subsection{Mixed Policy With Prior Knowledge Guidance}
The dual-arm planning problem in our work considers both the end-effectors' position and orientation. It is challenging to solve this high dimension planning problem if learning directly with a neural network since the explorable space is too large. By introducing the prior knowledge of specific problems, the explorable space can be limited to the vicinity of the optimal policy, thus improving the sample efficiency of reinforcement learning \cite{cheng2019control}. The mixed policy with prior knowledge can be defined as

\begin{equation}
\label{eq10}
\begin{array}{l}
{u_k}(s) = w{u_{\theta _k}}(s) + (1 - w)u_{\rm{prior}}(s)\\
s.t. \quad 0 \le w \le 1 
\end{array}
\end{equation}   
where $w$ and $1-w$ represent the weight of the reinforcement learning policy $u_{\theta_{k}}(s)$ and the prior policy $u_{prior}(s)$.

\noindent Define that $\pi_{prior}$ is the stochastic analogue to the deterministic prior policy $u_{prior}$, such that $\pi_{prior}(a|s) = \mathds{1}(a = u_{prior}(s))$ where $\mathds{1}$ is the indicator function.

\begin{theorem}
\label{mixed policy}
Let $D_{sub} = D_{TV}(\pi_{opt}, \pi_{prior})$ be the bias between the optimal policy and the prior policy, then the policy bias (i.e. $D_{TV}(\pi_{k}, \pi_{opt})$) can be bounded as follows,
\begin{equation}
\label{eq11}
\begin{array}{l}
{D_{TV}}({\pi _k},{\pi _{opt}}) \ge {D_{sub}} - w{D_{TV}}({\pi _{\theta k}},{\pi _{{\rm{prior}}}})\\
{D_{TV}}({\pi _k},{\pi _{opt}}) \le (1 - w){D_{sub}}\quad  {\rm{as\; k}} \to \infty 
\end{array}
\end{equation}
\end{theorem}

\noindent \textbf{Proof:} For proof see Appendix \ref{prftheo1}.

\begin{figure}[H]
  \centering
  \includegraphics[width=\hsize]{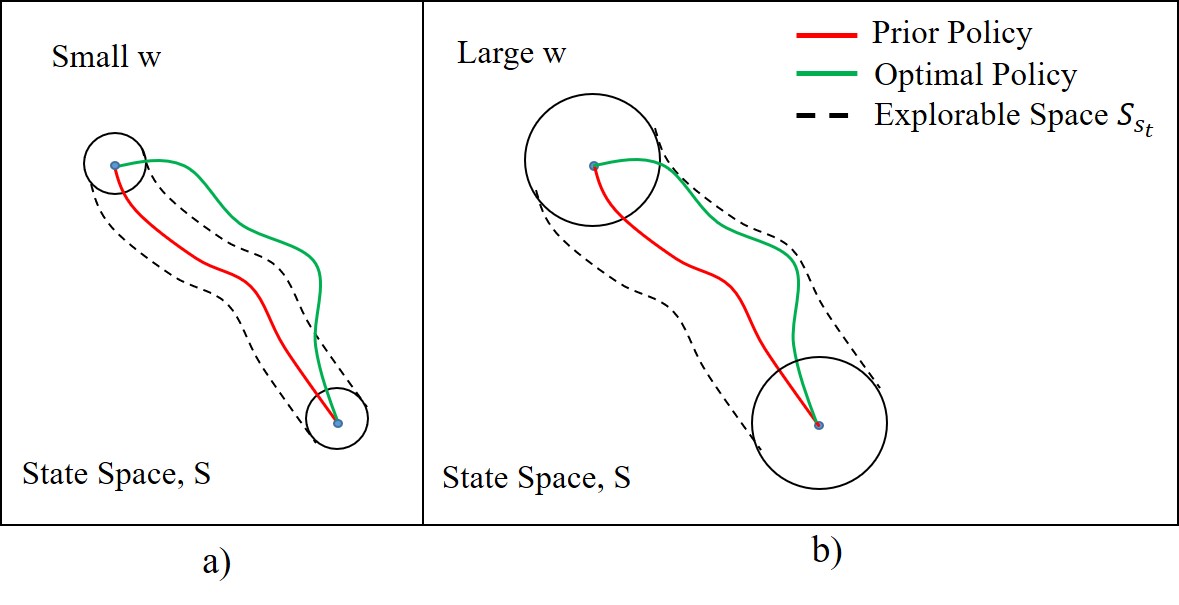}
  \caption{Illustration of the bound with the prior policy on the explorable space $S_{s_t}$. a) With small $w$, $S_{s_t}$ is too small to contain the optimal policy. b) With large w, $S_{s_t}$ is larger, so we can learn the optimal policy, but the explorable space is also enlarged.
}
  \label{Prior}
\end{figure}

\noindent The above theorem indicates that the bias between the mixed policy and the optimal policy is bounded by the prior bias during the training. Compared with an unbounded update, the mixed policy makes the early exploration more efficient. Fig. \ref{Prior} shows some intuition for the bounded consequence discussed above. The $\pi_{prior}$ limits the agent explorable space to $S_{s_{t}}$ near to the optimal policy. The region grows as $w$ increases and vice versa. In the case of $w = 1$, i.e., using pure reinforcement learning, the explorable space is no longer limited so that the agent needs to explore the whole state space in the training process resulting in low sample efficiency.

\noindent Thus, our research utilizes linear interpolation in Cartesian space as the prior policy of the planning algorithm. First, perform linear interpolation from the manipulator end-effectors to the target points. Then, calculate the joint angle errors between two adjacent interpolation points using inverse kinematics to obtain the desired joint angular velocities. However, due to the dynamic coupling, the inverse kinematics of the dual-arm free-floating space robot is complex. Besides, the manipulator will likely pass through the singular pose or reach the joint limits when utilizing inverse kinematics directly at the linear interpolation point.

\noindent Take two measures to solve the above problems. Firstly, use the inverse kinematics of the fixed-base manipulator to avoid complex modeling of the free-floating space robot. Secondly, when performing linear interpolation, we only consider the end-effector's position in Cartesian space. Any nonsingular orientation can be selected at the interpolation position to calculate joint angle errors. The desired joint angular velocities $\dot{\theta}_{t}$ output by the mixed policy is formulated as

\begin{equation}
\label{eq12}
{\hat{\dot{\theta_t}}} = w{a_t} + (1 - w){\dot \theta_{inv}}
\end{equation}
where $a_{t}$ is the output of the reinforcement learning Actor network, $w$ is its weight, and $\dot \theta_{inv}$ is the joint angle errors corresponding to adjacent interpolation points. To ensure the proportion of the two actions in the mixed policy depends on the weight $w$, both $a_{t}$ and $\dot \theta_{inv}$ range from -1 to 1.

\begin{figure}[H]
  \centering
  \includegraphics[width=\hsize]{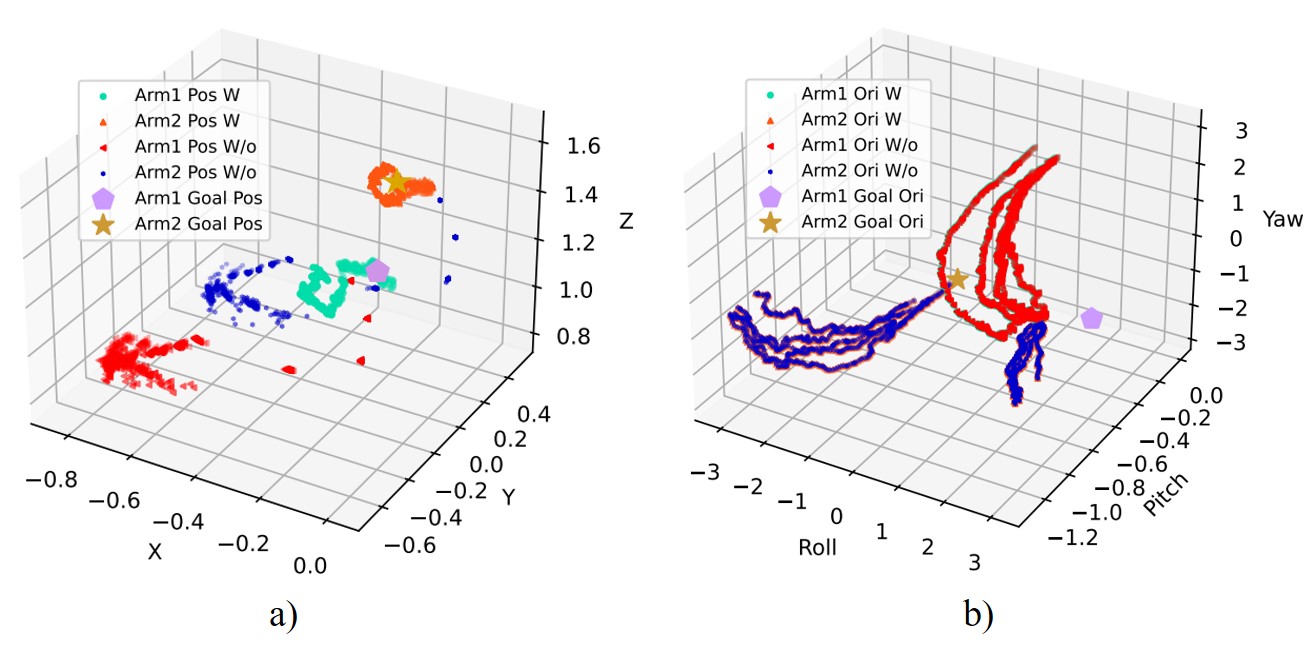}
  \caption{a): The position(XYZ) distribution of the end-effectors at the early training period with the mixed and pure learning policies. b): The orientation(RPY) distribution of the end-effectors at the early training period with the mixed and pure learning policies.}
  \label{fig4}
\end{figure}

\noindent Fig. \ref{fig4} shows the position and orientation distribution of the manipulator end-effectors at the early training period(the first five episodes) with the mixed policy as formulated in Eq. \ref{eq12} and the pure learning policy without prior knowledge guidance. The distribution of the end-effector positions is more concentrated near the target position with the mixed policy, while the orientation distribution is almost the same since no prior policy of planning orientation has been introduced. The sample distribution indicates that the linear interpolation of position facilitates samples with higher reward in the exploration, which solves the cold start problem caused by random exploration during the early training period and increases the sampling efficiency to expedite the algorithm's convergence.

\subsection{Efficient Reward Function}
In reinforcement learning, the reward function not only determines the optimization goal of the algorithm but also affects the convergence. This paper considers the manipulator's pose error from the end-effectors to the target and manipulability when designing the reward function.

\subsubsection{End-effector Error Reward Function}
In path planning, the position error between the end-effector and the target can be defined by the Euclidean distance $|d|$ between the two points. For reinforcement learning, the reward values are inversely proportional to the position error. Besides, considering that when the position error remains large, the reward needs to change sharply to expedite the convergence. At the same time, it is expected to be stable when the error is small to promote the end-effector maintaining near the target position. Therefore, $tanh$ is used to non-linearize $|d|$, and the reward function related to position error is defined as
\begin{equation}
{r_p} = (1 - \tanh (\left| d_1 \right|)) + (1 - \tanh (\left| d_2 \right|))
\end{equation}

\noindent In the dual-arm collaborative planning task considered in this paper, the position errors between the manipulator end-effectors and the target can be constrained to a small range due to the limitation of the arm length and the constraint of the closed-chain dual-arm workspace. In addition, the manipulator end-effector position changes more than the orientation under the same joint angle changes. Thus the end-effector position state space is small and can be explored efficiently. Compared with position error, the variation range of the end-effector orientation is $[-\pi, \pi]$, which is much larger than the end-effector position, resulting in low exploration efficiency in the orientation state space. Thus, a scalar reward function considering the three-axis orientation error can not efficiently guide the neural networks to converge to the optimal. Our work only considers the maximum error among the three-axis angles to calculate the reward at each time step. Employ the cosine function referring to the reward function design in \cite{lin2021collision}. The orientation reward function is defined as
\begin{equation}
{r_o} = \cos (\max e_{o_1}) + \cos (\max e_{o_2})
\end{equation}
$e_{o_i} = [{e_{r_i}},{e_{p_i}},{e_{y_i}}]$ represents error of roll, pitch and yaw.

\noindent The other three reward functions are compared with our design to prove that our method is more conducive to the algorithm's convergence. They are the $L1$ norm, $L2$ norm, and the average cosine values of the three-axis error. The expressions of the reward functions of one manipulator are
\begin{equation}
\begin{array}{l}
{r^i_{o1}} = 1 - \tanh (\left| {e_{r_i}} \right| + \left| {e_{p_i}} \right| + \left| {e_{y_i}} \right|)\\
{r^i_{o2}} = 1 - \tanh (\sqrt {{e_{r_i}}^2 + {e_{p_i}}^2 + {e_{y_i}}^2} )\\
{r^i_{o3}} = \frac{1}{3}(\cos ({e_{r_i}}) + \cos ({e_{p_i}}) + \cos ({e_{y_i}}))
\end{array}
\end{equation}

\noindent We can explain the benefit of our reward design from the perspective of gradient descent. As described in Eq. \ref{eq9}, the Actor network is updated along the gradient minimizing the orientation error to obtain the maximum reward in each episode. Since the three axes are symmetrical, for each reward function, fixed two-axis angle errors ${e_{p_i}}$ and ${e_{y_i}}$, the gradient of the remaining axis angle error can be derived as
\begin{equation}
\begin{array}{l}
{\dot r^i_o} =  - \sin (\max [{e_{r_i}},{e_{p_i}},{e_{y_i}}]) \\

{\dot r^i_{o1}} = {\tanh ^2}(\sqrt {{e_{r_i}}^2 + {e_{p_i}}^2 + {e_{y_i}}^2}  - 1),{\rm{          }}{{\rm{e}}_{r_i}} \ge 0\\

{\dot r^i_{o2}} = ({\tanh ^2}(\sqrt {{e_{r_i}}^2 + {e_{p_i}}^2 + {e_{y_i}}^2} ) - 1)\frac{{{e_{r_i}}}}{{\sqrt {{e_{r_i}}^2 + {e_{p_i}}^2 + {e_{y_i}}^2} }} \\

{\dot r^i_{o3}} =  - \sin ({e_{r_i}})
\end{array}
\end{equation}

\begin{figure}[H]
  \centering
  \includegraphics[width=\hsize]{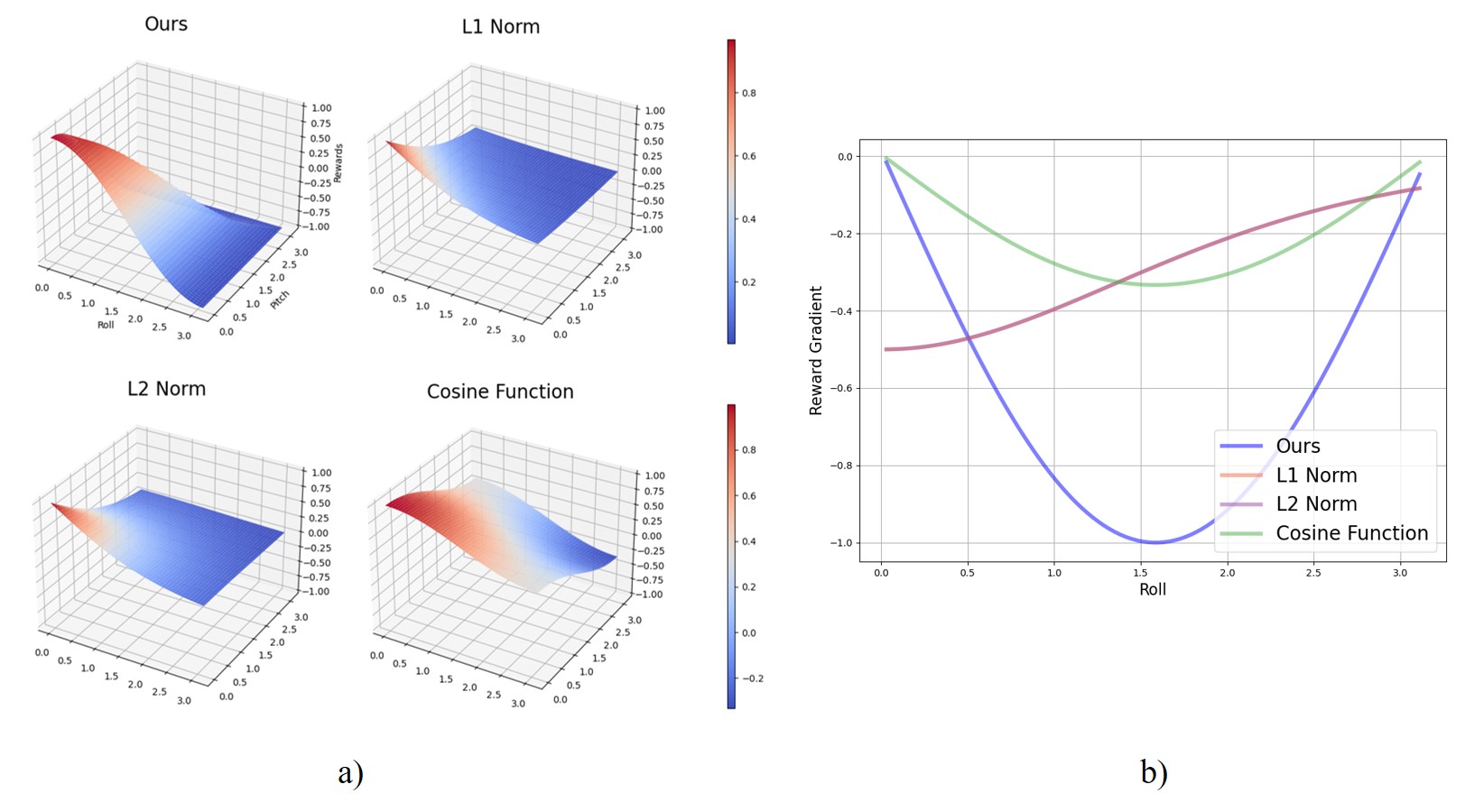}
  \caption{a): The function surfaces of four different reward functions with fixed yaw. b): The gradient curves of four different reward functions with fixed pitch and yaw.
  }
  \label{fig5}
\end{figure}

\noindent Fig. \ref{fig5} compares four different reward functions' function surfaces and gradient curves. It can be seen from Fig. \ref{fig5} a) that, for $L1$ norm and $L2$ norm, the agent obtains a low reward continuously due to the large error, which cannot provide instructive information to update the network. Fig. \ref{fig5} b) intuitively shows the gradients of the four reward functions on one-axis angle error. The update rate of the neural network is positively correlated with the gradient change rate of backpropagation. The gradient change rate of our designed reward function is the most rapid, so that can improve the convergence efficiency. 

\subsubsection{Manipulability Optimization}
The manipulability reflects the manipulator singularity under a specific configuration. Since the same target pose in Cartesian space corresponds to multiple configurations, the farther the planned path from the singularity, the smoother the manipulator motion will be. Refer to \cite{yoshikawa1985manipulability}, the manipulability of the manipulator is defined as
\begin{equation}
{M_m} = \sqrt {\det ({J}(\Theta ){{J}^{\mathop{\rm T}\nolimits} }(\Theta ))}
\end{equation}
$\Theta$ is the joint angles of the manipulator in the current configuration, and $J$ is the corresponding Jacobian matrix. 

\noindent Then, the reward function to optimize the manipulability of the manipulators in operation can be formulated as
\begin{equation}
r_{m} = {w_{1}}\sqrt {\det ({{J}_1}{J}_1^{\mathop{\rm T}\nolimits} )}  + {w_{2}}\sqrt {\det ({{J}_2}{J}_2^{\mathop{\rm T}\nolimits} )}
\end{equation}
$w_{i}$ represents the weight of manipulability of each manipulator. In our work, the two manipulators are treated equally, thus$ w_{1} = w_{2} = 0.5$.

\noindent It is specified that when the position error is less than $0.02m$ and the maximum axis angle error is less than $0.05rad$, the planning task is successful. Then set the reward to be 3. Therefore, the final reward design is
\begin{equation}
\label{eq19}
r = \left\{ 
\begin{aligned}
r_{p} + r_{o} + r_{m}, \quad &others \\
3 \quad, \quad &|d_{i}| \leq 0.02 \ and \ \max {e_{o_i}} \leq 0.05
\end{aligned}
\right.
\end{equation}

\noindent For the pseudo code of the algorithm proposed in this paper, see Appendix \ref{Pseudo-code}.

\section{Simulation Results}
\subsection{Simulation Environment}
The simulation environment for the capture operation of the 6-DOF dual-arm free-floating space robot is built in Mujoco, a typical robot physical simulation platform, as shown in Fig. \ref{fig6}. The base satellite's parameters and the manipulator's joint limitations are shown in Appendix \ref{envdesign}. The movement of the manipulators will interfere with the posture of the uncontrolled base satellite due to the dynamic coupling, which in turn affects the end-effectors' precision. Collision detection is carried out in the whole training process, including the collision between the components of the system and the collision with the target object in space.

\begin{figure}[H]
  \centering
  \includegraphics[width=\hsize]{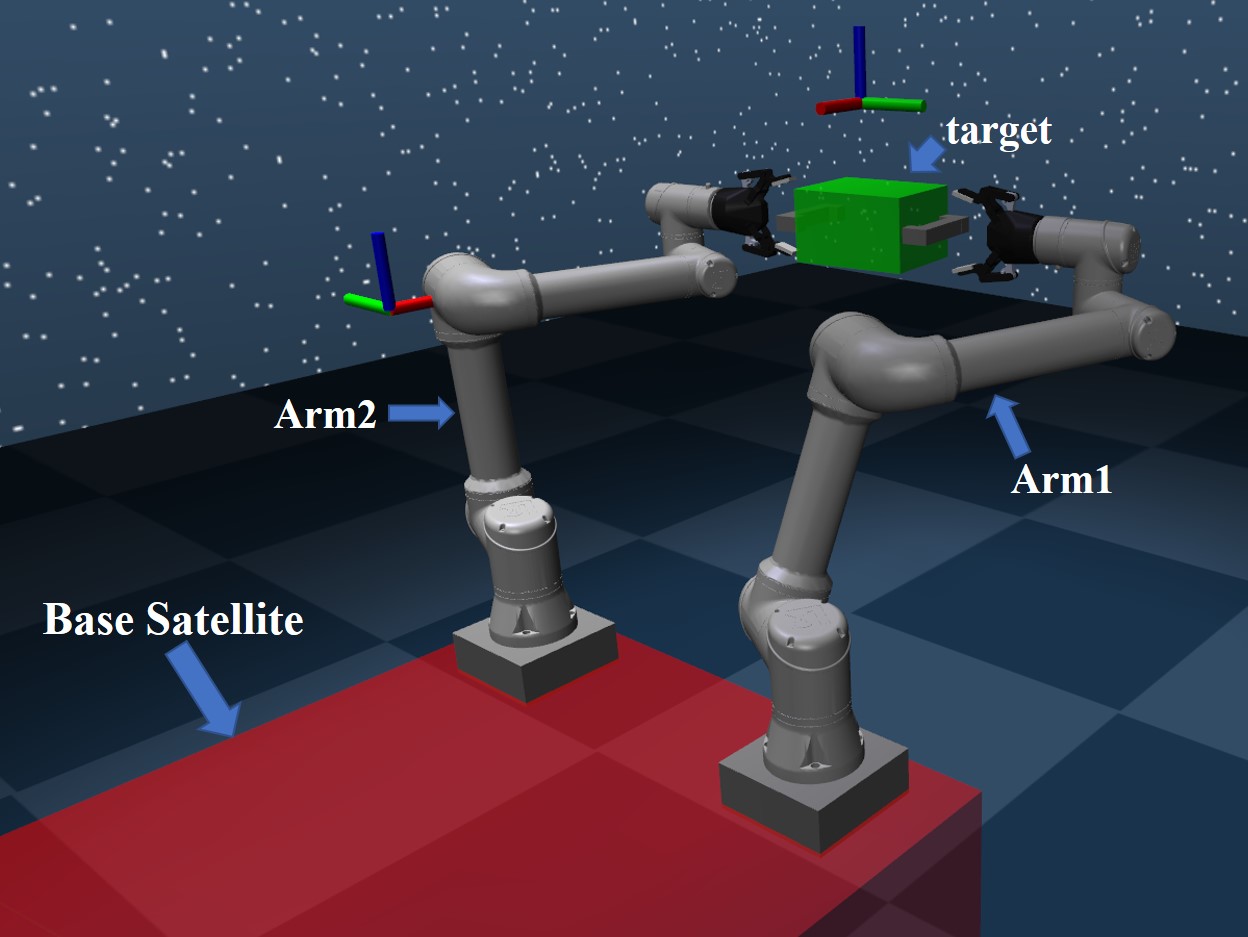}
  \caption{Simulation environment of the dual-arm free-floating space robot capture operation.}
  \label{fig6}
\end{figure}

\noindent The planning problem discussed in this paper takes the position and orientation errors between the end-effectors and the target as the evaluation index of the algorithm performance. The force control after contact between the end-effectors and the target is not involved. The task is to plan a collision-free trajectory that enables the end-effectors to reach the position and orientation of the target rotating about a single axis and to maintain stable tracking within the dual-arm operable workspace. Therefore, in the training phase, we fixed the target's center position and randomly set the target orientation along the z-axis(target coordinate) between $[-\frac{\pi}{3}, \frac{\pi}{3}]$. The target remains stationary unless a collision occurs in each episode.

\noindent For the training environment of reinforcement learning and the setting of algorithm hyperparametric, see Appendix \ref{envdesign}.

\subsection{Performance Comparison}
\begin{figure}[H]
  \centering
  \includegraphics[width=\hsize]{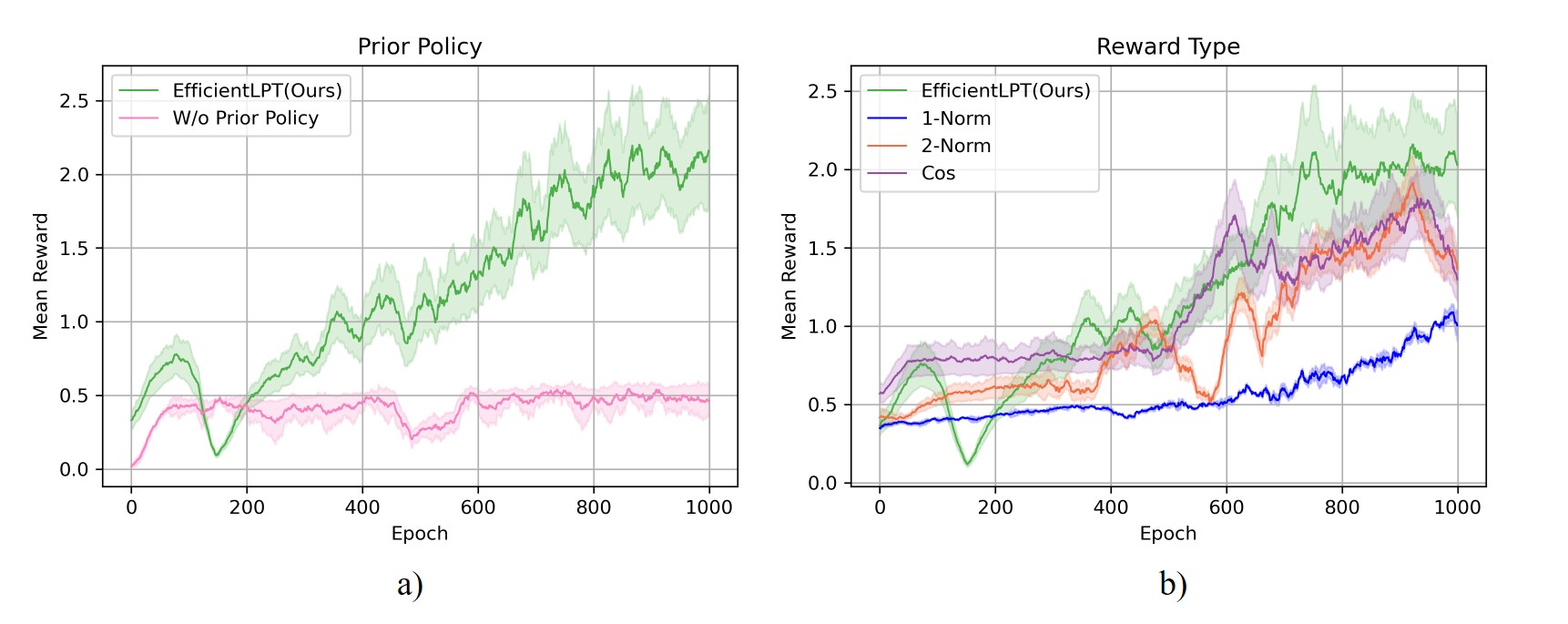}
  \caption{a): The mean reward curve in the training process of our method and pure reinforcement learning policy. b): The mean reward curve in the training process with different reward functions under the condition of using the prior policy.}
  \label{fig7}
\end{figure}

Fig. \ref{fig7} a) compares the mean reward curve in the training process of our method with the prior policy and pure reinforcement learning policy, wherein each epoch includes five episodes. Set the random seeds to 0, 20, and 100. Although the mean reward of the pure reinforcement learning algorithm rises steadily in the first 250 epochs, it rises very slowly, and the convergence is unstable. In contrast, the algorithm combining a prior policy fluctuates sharply in the early stage of training. This may be because the agent needs to learn to integrate the existing prior policy and newly learned policy in the early period. In addition, the infinite norm of orientation error in the reward function also exacerbates the fluctuation. After 150 epochs, the mean reward obtained by the algorithm with prior policy increases steadily and tends to be stable after 800 epochs. The convergence of our method is significantly more stable than learning directly from the initial random policy.  

\noindent Fig. \ref{fig7} b) shows the mean reward curve in the training process with different reward functions using the prior policy. The convergence stability using $L1$ norm, $L2$ norm, and cosine function is better than our designed reward function before 200 epochs. However, after that, the mean reward of our method rises fastest among the four reward functions. The mean reward curve of the $L1$ norm rises most stably, but the convergence speed is the slowest. Compared with the $L2$ norm and the cosine function, the mean reward curve of our method has higher stability. It is verified that the reward function adopted in this paper improves the algorithm's convergence efficiency and maintains high stability.

\noindent Table \ref{success rate} shows the success rates of different algorithms in reaching random targets in the test environment after training 1000 epochs. Since the position and orientation planning of dual arms is a complex problem with high dimensions, the prior policy of position inverse kinematics is significant in solving this problem. In the test environment, the algorithm's success rate without prior policy is 0. Furthermore, designing an efficient reward function also impacts the algorithm's performance. The success rate of our designed reward function is much higher than others.

\begin{table}[h]
    \centering
    \caption{Test success rate of different algorithms.}
    \begin{tabular}{c|c}
    \toprule  
    Algorithms & Success rate \\\hline
    \textbf{EfficientLPT(ours)} & 100 $\%$ \\
    Pure RL & 0 $\%$ \\
    $L1$ Norm &	56 $\%$ \\
    $L2$ Norm &	59 $\%$ \\
    Cosine function & 40 $\%$ \\
    \bottomrule 
    \end{tabular}
    \label{success rate}
\end{table}

\begin{figure}[H]
  \centering
  \includegraphics[width=\hsize]{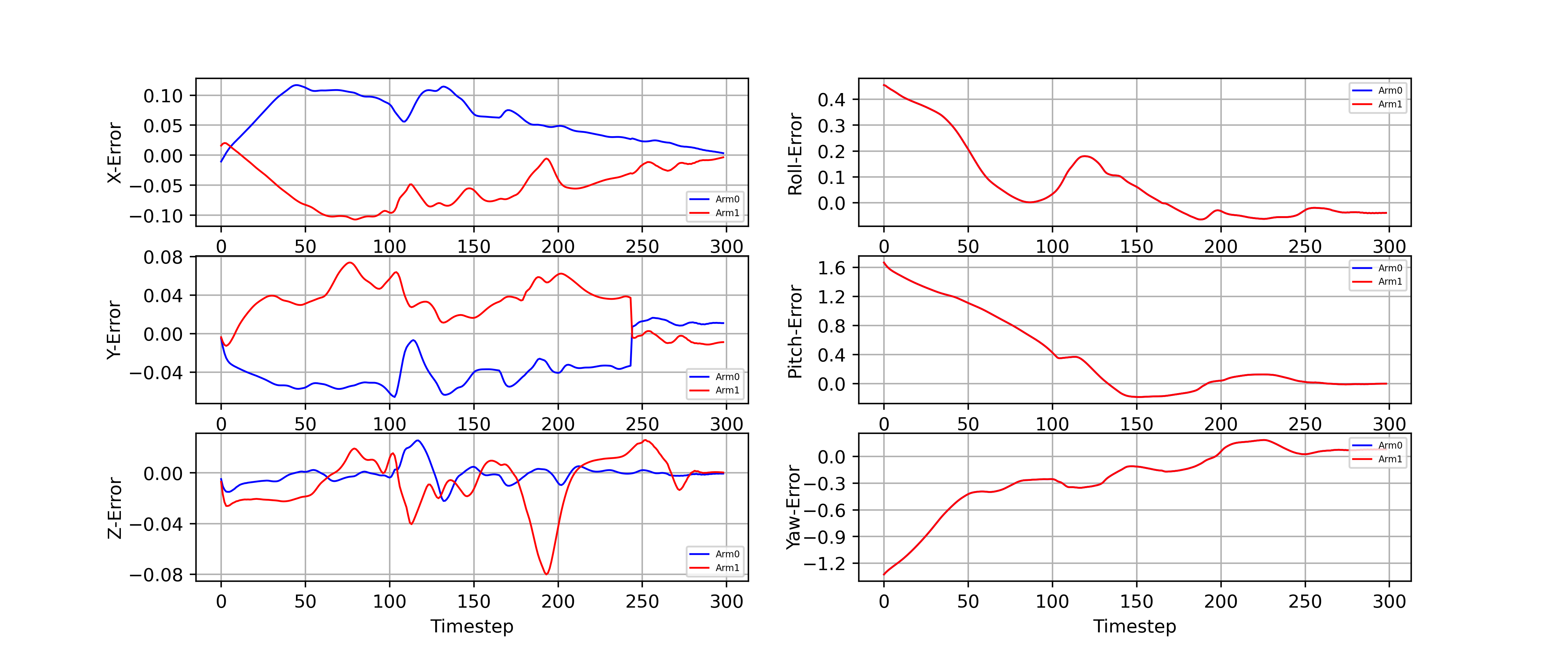}
  \caption{The position error and orientation error between the end-effectors and the target in the reaching process to a stationary target.}
  \label{fig8}
\end{figure}

\noindent Fig. \ref{fig8} shows the position and orientation error between the end-effectors and the target in the manipulator reaching process to a stationary target. The target position is $[ -0.1012, -0.0849, 1.5094]$ in the world coordinate and the target orientation is $[-0.49345, 0, 0]$. The dual-arm end-effectors can simultaneously reach the target within an acceptable error range, which indicates that the planning method can achieve accurate path planning to reach a stationary target.

\subsection{Tracking of Rotating Target}
\begin{figure}[H]
  \centering
  \includegraphics[width=\hsize]{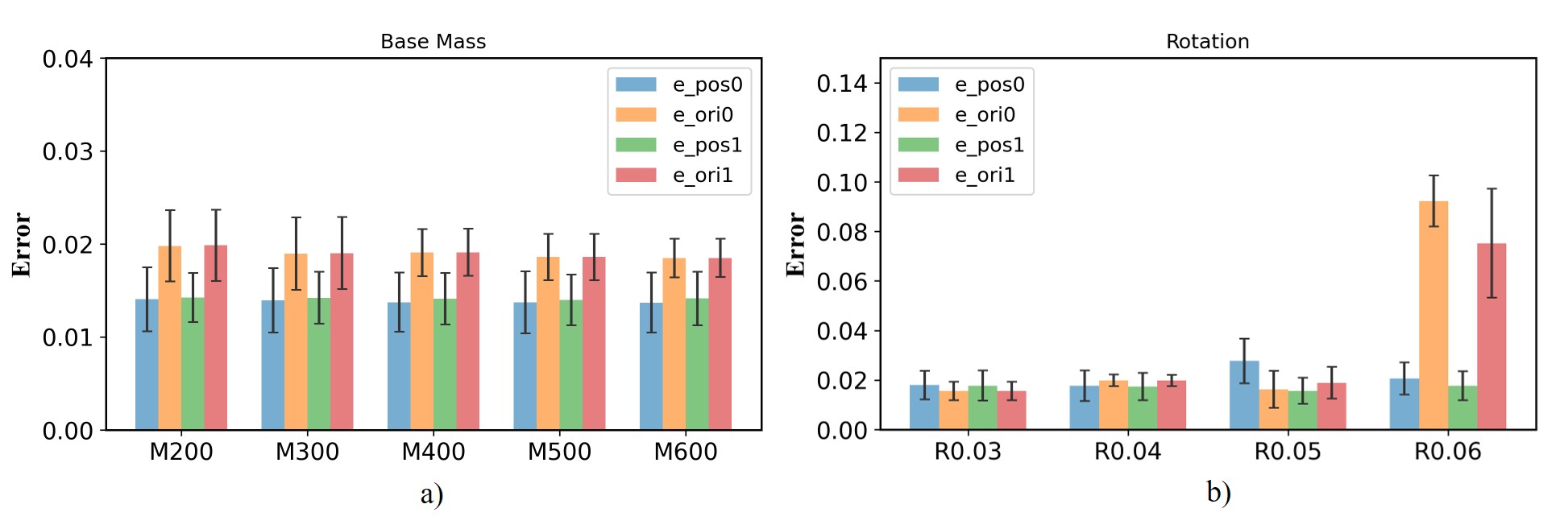}
  \caption{a): The position and orientation error in 20 episodes(stationary target with random orientation) with different base satellite masses. b): The mean and standard deviation of the position and orientation error of the end-effectors tracking the target with different rotating speeds.}
  \label{fig9}
\end{figure}

For the dual-arm free-floating space robot discussed in this paper, the disturbance to the base satellite posture during the manipulator movement is slight due to the large mass of the base satellite. However, in tasks requiring high accuracy, like reaching path planning, the interference caused by the base motion can not be ignored. To verify the robustness of our method facing base motion disturbance, we test the position and orientation error in 20 episodes with different base satellite masses(stationary target with random orientation), as shown in Fig. \ref{fig9} a). The position errors are all less than $0.02m$, and the orientation errors are less than $0.05rad$, which indicates that the algorithm can effectively compensate for the interference caused by the base satellite motion.

\noindent Test the planning algorithm proposed in this paper to reach the target rotating along the z-axis(in target coordinate). The rotating speed is set from $1.72\,^{\circ}/s$ to $3.44^{\circ}/s$. Fig. \ref{fig9} b) shows the mean and standard deviation of the position error and orientation error of the end-effectors tracking target, i.e., the pose error of each step after the position error is less than $0.02m$ and the orientation error is less than $0.05rad$. When the target's self-rotating speed is slow, the manipulators can reach the target without predicting its motion and realize continuous and stable tracking, which verifies the effectiveness and robustness of our method in planning and tracking the moving target. However, when the self-rotating speed reaches $3.44^{\circ}/s$, although the position error is still less than $0.02m$, the orientation error is close to $0.1rad$, which can not meet the trajectory tracking error requirements. The reason is that to avoid the system instability caused by the too-violent motion of the manipulators, we strictly limit joint angular velocities to a small range. Our motion planning method is suitable for tracking and capturing slowly rotating targets. For space operations, direct contact between the space robot and a target with enormous momentum may cause external force beyond the range, causing damage to the manipulators \cite{wu2018contact}. Generally, the target will be de-tumbled first to reduce the relative velocity between the end-effector and the target \cite{gang2020detumbling}. Therefore, the method proposed in this paper applies to the actual scene of space operation.

\begin{figure}[H]
  \centering
  \includegraphics[width=\hsize]{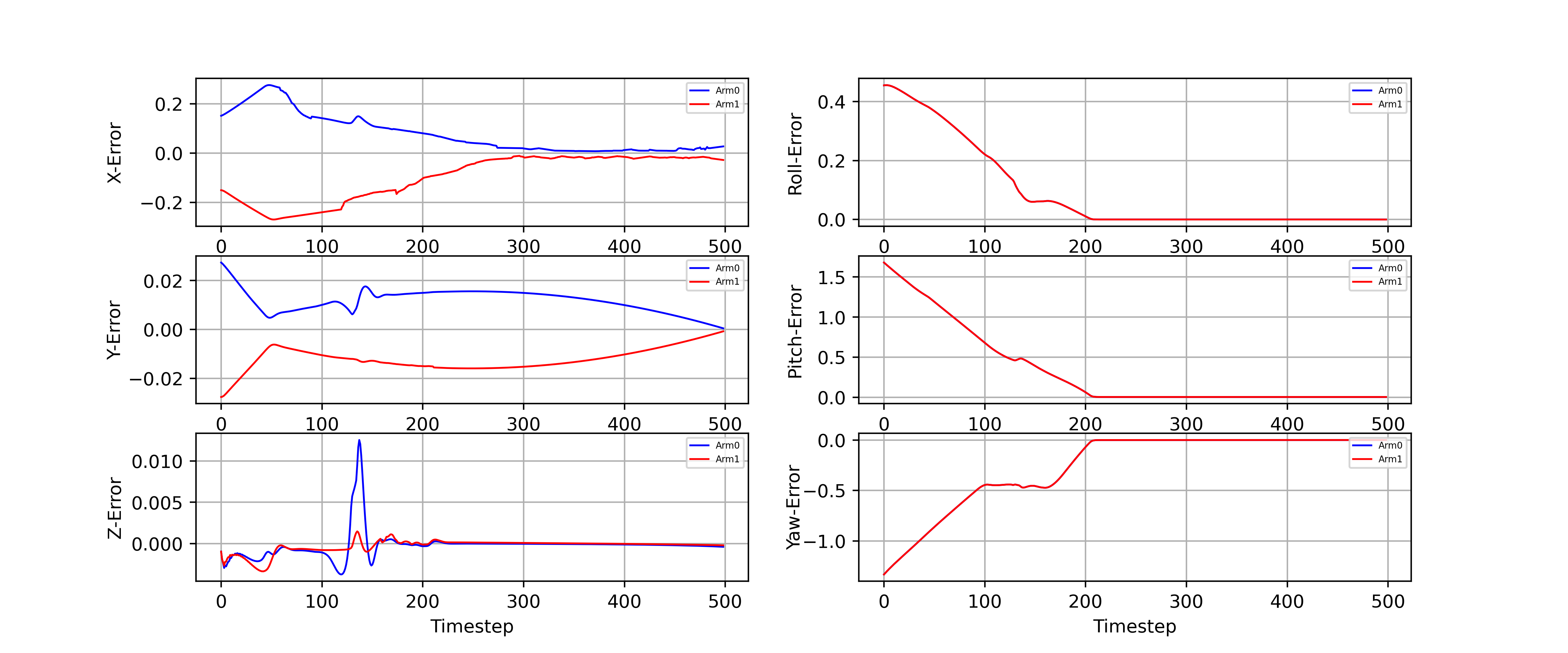}
  \caption{The position and orientation error of the manipulators.}
  \label{fig10}
\end{figure}

\begin{figure}[H]
  \centering
  \includegraphics[width=\hsize]{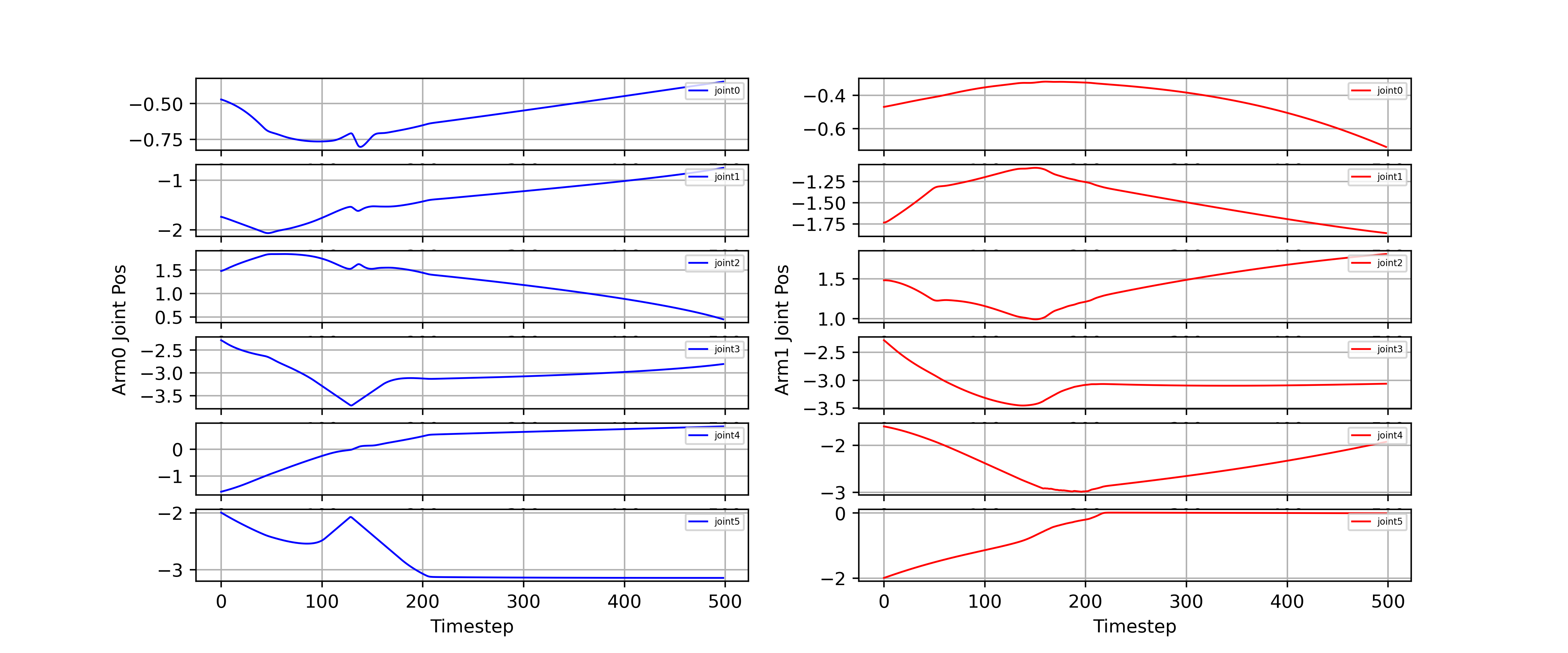}
  \caption{The manipulators joint angles.}
  \label{fig11}
\end{figure}

\begin{figure}[H]
  \centering
  \includegraphics[width=\hsize]{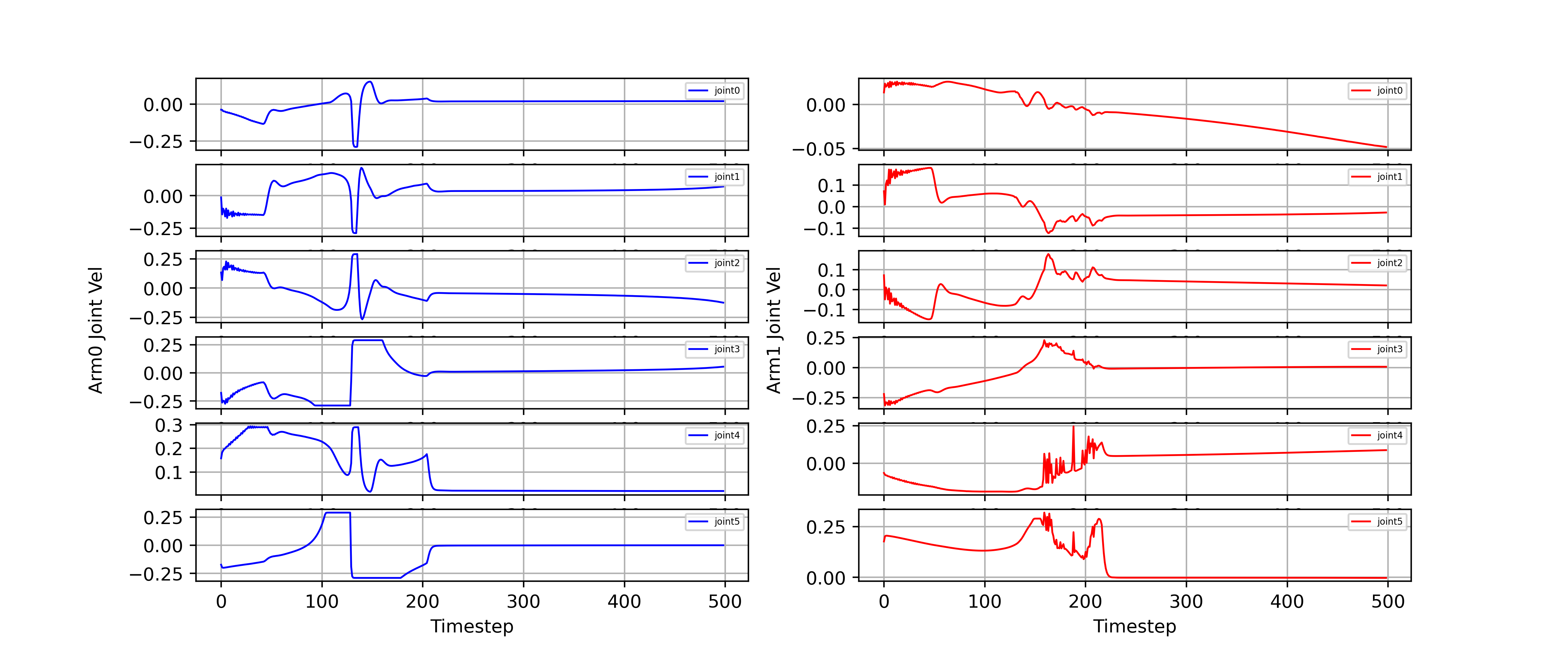}
  \caption{The manipulators joint angular velocities.}
  \label{fig12}
\end{figure}

\noindent Fig. \ref{fig10} to \ref{fig12} shows the pose error, the joint angles, and the joint angular velocities of the manipulators when reaching the target with a rotating speed of $2.29^{\circ}/s$. Fig. \ref{fig10} shows that the manipulators reach the target pose at about 300 steps and keep accurate tracking after that. It can be seen from Fig. \ref{fig11} and \ref{fig12} that the joint angles and angular velocities are continuous during the tracking, which indicates that the actions generated by the mixed policy are smooth, providing a basis for the later sim-to-real experiment.

\begin{figure}[H]
  \centering
  \includegraphics[width=\hsize]{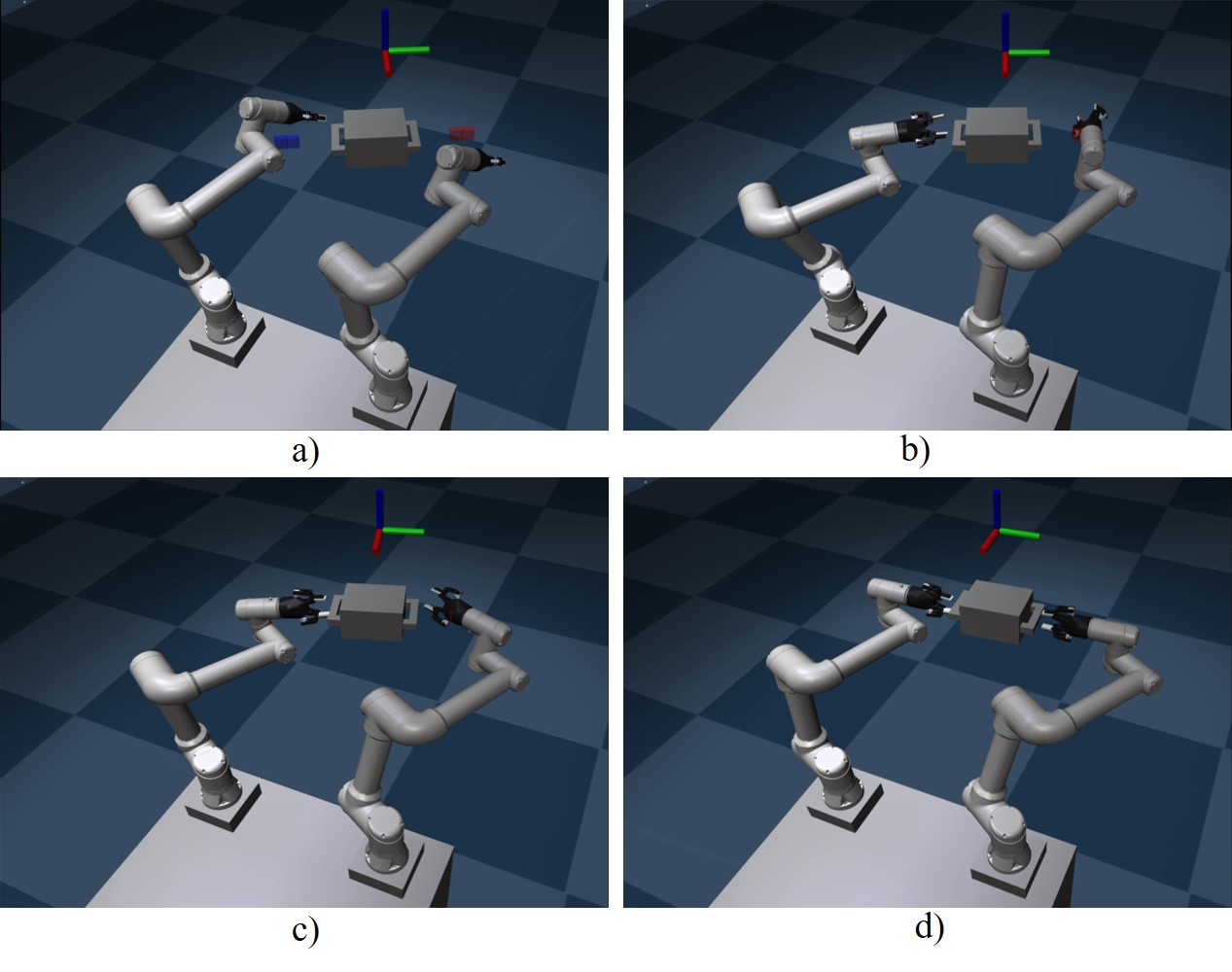}
  \caption{Reaching and tracking process of dual-arm space robot for a rotating target.}
  \label{fig13}
\end{figure}

\noindent In order to intuitively demonstrate the process of the dual-arm free-floating space robot reaching and tracking the slow-rotating target, Fig. \ref{fig13} shows the screenshots of the movement process at several representative time nodes.

\section{Conclusion}
For the current algorithms, some critical problems remain unsolved for the dual-arm free-floating space robot planning problem. On the one hand, the dynamic coupling of the free-floating space robot system will affect the end-effector's motion accuracy. On the other hand, the offline planning/online tracking algorithms cannot solve the problem of planning and tracking uncertain moving targets. We propose a motion planning method for dual-arm free-floating space robots based on reinforcement learning. The problem of slow and unstable convergence of reinforcement learning in high-dimensional planning problems is solved by introducing the prior policy of the manipulator inverse kinematics and the infinite norm of the orientation error of the reward function.

\noindent The simulation results show that the proposed algorithm has strong robustness and can realize the planning and tracking of the target with different spin speeds with a short decision-making time and a high success rate so that to meet the real-time application requirements. In the future, we will consider planning and tracking tumbling targets with a more extensive motion range in position and orientation.

\bibliographystyle{unsrt}  
\bibliography{main} 

\appendix

\section*{Appendix}
\section{Details of Theoretical Analysis}
\subsection{Proof of Theorem 1}
\label{prftheo1}

\begin{proof}
Let us define the policy bias as $D_{TV}(\pi_k, \pi_{opt})$, and $D_{sub} = D_{TV}(\pi_{opt}, \pi_{prior})$. Since $D_{TV}$ is a metric that represents the total variational distance, we can use the triangle inequality to obtain:
\begin{equation}
\label{eq20}
{D_{TV}}({\pi _k},{\pi _{opt}}) \ge {D_{TV}}({\pi_{prior}},{\pi _{opt}}) - {D_{TV}}({\pi_{prior}},{\pi _k})
\end{equation}
According to the mixed policy definition in Eq. \ref{eq10}, we can further decompose the term ${D_{TV}}({\pi_{prior}},{\pi _k})$:
\begin{equation}
\label{eq21}
\begin{aligned}
{D_{TV}}({\pi _{prior}},{\pi _k}) &= \mathop {\sup }\limits_{(s,a) \in S{\rm{x}}A} \left| {{\pi _{prior}} - w{\pi _{{\theta _k}}} - (1 - w){\pi _{prior}}} \right|\\
 &= w\mathop {\sup }\limits_{(s,a) \in S{\rm{x}}A} \left| {{\pi _{{\theta _k}}} - {\pi _{prior}}} \right|\\
 &= w{D_{TV}}({\pi _{{\theta _k}}},{\pi _{prior}})
\end{aligned}
\end{equation}

\noindent This holds for all $k \in \mathbb{N}$ from Eq. \ref{eq20} and Eq. \ref{eq21}, and we can obtain the lower bound in Eq. \ref{eq22},
\begin{equation}
\label{eq22}
{D_{TV}}({\pi _k},{\pi _{opt}}) \ge {D_{sub}} - w{D_{TV}}({\pi _{\theta k}},{\pi _{{\rm{prior}}}})
\end{equation}

\noindent To obtain the upper bound, since the RL policy $\pi_{\theta_{k}}$ can achieve asymptotic convergence to the (locally) optimal policy $\pi_{opt}$ through the policy gradient algorithm (as proven for certain classes of function approximators in \cite{1999Policy}), denote this policy as $\pi^{(p)}_{\theta_{k}}$, such that $\pi^{(p)}_{\theta_{k}} \to \pi_{opt}$ as $ k \to \infty$. In this case, we can derive the bias between the mixed policy $\pi_k$ and the optimal policy as follows,
\begin{equation}
\label{}
\begin{aligned}
{D_{TV}}({\pi _{opt}},\pi _k^{(p)}) &= \mathop {\sup }\limits_{(s,a) \in S{\rm{x}}A} \left| {{\pi _{opt}} - w\pi _{{\theta _k}}^{(p)} - (1 - w){\pi _{prior}}} \right|\\
 &= (1 - w)\mathop {\sup }\limits_{(s,a) \in S{\rm{x}}A} \left| {{\pi _{opt}} - {\pi _{prior}}} \right| \quad \quad  {\rm{as\; k}} \to \infty \\
 &= (1 - w){D_{TV}}({\pi _{opt}},{\pi _{prior}}) \quad  {\rm{as\; k}} \to \infty  \\
 &= (1 - w){D_{sub}}\quad  {\rm{as\; k}} \to \infty 
\end{aligned}
\end{equation}

\noindent Therefore, we can obtain the upper bound:
\begin{equation}
\label{}
\begin{aligned}
{D_{TV}}({\pi _k},{\pi _{opt}}) &\le {D_{TV}}({\pi^{(p)} _k},{\pi _{opt}}) \\ 
&=(1 - w){D_{sub}}\quad  {\rm{as\; k}} \to \infty 
\end{aligned}
\end{equation}

\end{proof}

\section{Environmental Design}
\label{envdesign}
The parameters of the base satellite and the joint limitations of the manipulator are shown in Table \ref{tab1}. The base satellite is equipped with two 6-Dof UR5 manipulators whose kinematics and dynamics parameters are the official default values.

\begin{table}[!htb]
  \centering
     \caption{Parameters of dual-arm free-floating space robot}
    \begin{tabular}{c|c|c}
    \toprule  
    Component & Parameters & Values \\\hline
      & Mass[kg] & 419.8441 \\
    Base satellite & Size[m] & (0.5326, 0.5326, 0.3) \\
      &	Offset of the manipulator base & (0.37, $\pm 0.34$, 0.3726) \\\hline
    Manipulators & Joint angle limitation[rad] & $\pm 6.28319$, $\pm 3.14159$ \\
      &	Joint torque limitation[Nm] & $\pm 150$, $\pm 28$ \\
    \bottomrule 
    \end{tabular}

  \label{tab1}
\end{table}

\section{Details of Algorithms}
\subsection{Training Environment and Hyperparameters}
\label{Training}
The simulation is carried out with Python 3.7, and the SAC algorithm is achieved using the Pytoch 1.1.0. The hyperparameters of \textbf{SAC} is illustrated in Table \ref{hp of sac}. 

\begin{table}[h]
    \centering
    \caption{Hyperparameters of \textbf{SAC}}
    \begin{tabular}{c|c}
    \toprule  
    Hyperparameters& SAC\\\hline
    Actor network & (256,256) \\
    Critic network & (256,256) \\
    Learning rate of actor & 1.e-3  \\
    Learning rate of critic & 5.e-4  \\
    Optimizer & Adam \\
    ReplayBuffer size& $10^6$ \\
    Discount ($\gamma$)& 0.995 \\
    Polyak ($1-\tau$)& 0.995 \\
    Batch size & 128 \\
    Length of an episode & 200 steps \\
    Maximum steps & 1e6 steps \\
    \bottomrule 
    \end{tabular}
    \label{hp of sac}
\end{table}

\subsection{ Pseudo-code of Algorithm}
\label{Pseudo-code}

The pseudo-code of the EfficientLPT algorithm proposed in this paper is as follows.

\begin{algorithm}[]
	\renewcommand{\algorithmicrequire}{\textbf{Input:}}
	\renewcommand{\algorithmicensure}{\textbf{Output:}}
	\caption{ Efficient Learning-based Path Tracking(EfficientLPT)}
	\label{alg:1}
	\begin{algorithmic}[1]
	    \STATE Randomly initialize the parameters of actor and two critic networks with $\phi,\theta_{1}, \theta_{2}$
	    \STATE Initialize replay buffer $\mathcal D$ and $w$
	    \STATE Initialize the parameters of target network with $\theta^\prime_{i} \leftarrow \theta_{i}, i=1,2$
	    \FOR {episode $m = 1, M $}
	     \STATE Sample initial state $s_0$ 
	    \FOR {step $t=0, T-1$} 
	    \STATE Calculate the position error $e_p$ and orientation error $e_o$ between the current end-effectors and the target
	    \STATE Linearly interpolated $e_p$ and $e_o$, and obtained the interpolation point $X_{e}$ closest to the current end-effector pose
	    \STATE Calculate $\dot{\theta}_{inv}$ through the error between the joint angles corresponding to $X_{e}$ and the current joint angles
	    \STATE Sample a action $a_t$ from $\pi_\phi(a_t|s_t)$
	    \STATE Caculate the desired joint angular velocities $\hat{\dot{\theta_t}}$ according to Eq.\ref{eq12}
	    \STATE Execute the mixed action $\hat{\dot{\theta_t}}$ and observe a new state $s_{t+1}$
	    \STATE Store $<s_t,a_t,r_t,s_{t+1}>$ into $\mathcal{D}$
	    \ENDFOR
	    \FOR{iteration $n = 1, N$}
	    \STATE Sample a minibatch $\mathcal B$ from the replay buffer $\mathcal D$
	    \STATE Update the $\theta_{i}, i=1,2$ according to Eq.\ref{eq7} using minibatch $\mathcal B$
	    \STATE Update the $\phi$ according to      
             Eq.\ref{eq9} using minibatch $\mathcal B$
	    \STATE Update the parameters of target networks, $\theta^\prime_{i}, i=1,2$
	    \ENDFOR
	    \ENDFOR
	\end{algorithmic}
\end{algorithm}

\end{document}